%% file: closer.tex
\documentclass[a4paper,twoside]{article}

\usepackage{times}
\usepackage{latexsym}
\usepackage{verbatim}
\usepackage{bm}
\usepackage{amsmath}
\usepackage{graphicx}
\usepackage{latexsym}
\usepackage{booktabs}
\usepackage{multirow}
\usepackage{subcaption}
\usepackage{epstopdf}
\usepackage{multirow}
\usepackage{footnote}
\usepackage{tablefootnote}

\usepackage{hyperref}
\usepackage{algorithm}
\usepackage{algpseudocode}
\usepackage[table]{xcolor}
\usepackage[textsize=tiny,color=yellow]{todonotes}

\usepackage{SCITEPRESS}

\def \ie {\emph{i.e.\ }}
\def \eg {\emph{e.g.\ }}

\begin{document}
\title{A Deep Learning based approach\\ to VM behavior identification in cloud systems}
\author{
\authorname{Matteo Stefanini, Riccardo Lancellotti, Lorenzo Baraldi, Simone Calderara}
\affiliation{Department of Engineering "Enzo Ferrari", University of Modena and Reggio Emilia, Modena, Italy}
\email{\{name.surname\}@unimore.it}
}

\abstract{
Cloud computing data centers are growing in size and complexity to the point where monitoring and management of the infrastructure become a challenge due to scalability issues. A possible approach to cope with the size of such data centers is to identify VMs exhibiting a similar behavior. Existing literature demonstrated that clustering together VMs that show a similar behavior may improve the scalability of both monitoring and management of a data center. However, available clustering techniques suffer from a trade-off between the accuracy of the clustering and the time to achieve this result. Not being able to obtain an accurate clustering in short time hinders the application of these solutions, especially in public cloud scenarios where on-demand VMs are instantiated and run for a short time span.
Throughout this paper we propose a different approach where, instead of an unsupervised clustering, we rely on classifiers based on deep learning techniques to assign a newly deployed VMs to a cluster of already-known VMs. The two proposed classifiers, namely DeepConv and DeepFFT use a convolution neural network and (in the latter model) exploits Fast Fourier Transformation to classify the VMs.
Our proposal is validated using a set of traces describing the behavior of VMs from a real cloud data center. The experiments compare our proposal with state-of-the-art solutions available in literature, such as the AGATE technique and PCA-based clustering, demonstrating that our proposal can achieve a very high accuracy (compared to the best performing alternatives) without the need to introduce the notion of a gray-area to take into account not-yet assigned VMs as in AGATE. Furthermore, we show that our solution is significantly faster than the alternatives as it can produce a perfect classification even with just a few samples of data, such as 4 observations (corresponding to 20 minutes of data), making our proposal viable also to classify on-demand VMs that are characterized by a short life span.
}
\onecolumn \maketitle \normalsize \vfill

\input{1-intro.tex}
\input{2-model.tex}

\input{3-results.tex}

\input{4-related.tex}
\input{5-conclusions.tex}

\bibliographystyle{IEEEtran}
\bibliography{cloud}
\end{document}

%% file: 1-intro.tex
\section{Introduction}\label{sec:intro}

The popularity of cloud computing is clearly demonstrated by its wide adoption: for example, nearly 60\% of the Apache Spark installations are deployed in the Cloud~\cite{cloudera2018}. The benefit from embracing the Cloud paradigms typically lies in the reduced cost of ownership for the infrastructure, that may reach up to 66\% for some cases~\cite{aws2012}.

A critical point for the IaaS Cloud infrastructures is the monitoring and management of the virtual machines (VMs) and physical nodes in a data center. Even just monitoring may present scalability issues, due to the sheer amount of data involved~\cite{NRDC:2014}. In a similar way, the optimization problem for mapping VMs over the infrastructure may exhibit an unmanageable dimensionality forcing the Cloud provider to introduce some over-simplification~\cite{Katz}. A common problem that hinders the scalability of monitoring and management in Cloud data centers is considering each VM as a black-box independent from the others.
Effective proposal to improve the scalability of Cloud monitoring and management using a class based-approach have been recently introduced~\cite{Canali:TGCN, Canali:NCCA15}. These class-based solutions leverage the observation that VMs hosting the same software component of the same application exhibit similar behavior with respect to resource utilization. Hence, by taking into account the similarity in VMs behavior it is possible, for example, to increase by nearly one order of magnitude the number of VMs that can be considered in the of the data-center VMs allocation problem \cite{Canali:NCCA15}. 
However, available clustering techniques to identify VMs that exhibit similar behavior, show a trade-off between accuracy of VMs group identification and the amount of observations (and hence the time) required to reach an accurate classification~\cite{Canali:TCC18}. This issue hinders the application of a class-based approach outside a static scenario characterized by long-term commitments~\cite{Durkee:2010}, where cloud customers purchase VMs for extended periods of time. An attempt to address this trade-off has been made through a technique named AGATE (Adaptive Gray Area-based TEchnique)~\cite{Canali:TCC18}, that adapts the amount of observations to the level of certainty of the identification of a VM as belonging to a cluster. However, also this technique does not ensure an upper bound on the time to identify VMs.  

In this paper we propose a different approach based on a classifier that uses a deep learning technique to identify the VMs relying on a model obtained from preliminary training. While the need to tune the neural network reduces the flexibility compared to the purely clustering-based approach proposed in~\cite{Canali:TCC18}, the proposed classifier is significantly faster and can identify VMs with a perfect accuracy, observing their behavior for just a few minutes, compared to the hours required in the alternative approaches.

We tested the proposed classifier using traces derived from the work in~\cite{Canali:TCC18}. The results confirm that our proposal significantly outperforms the alternatives in terms of accuracy in the classification and in the time required to reach that accuracy.


The remainder of this paper is organized as follows. 
Section~\ref{sec:problem} provides a description of the proposed deep-learning-based classifier including both the model and the implementation.
Section~\ref{sec:results} describes the experimental results. Section~\ref{sec:related} discusses the related work and  Section~\ref{sec:conclusions} concludes the paper with some final remarks and outlines future research directions.


%% file: 2-model.tex
\section{Methodology}\label{sec:problem}
Our contribution comprises two learnable models for VM identification, which we call \textit{DeepConv} and \textit{DeepFFT}. While the former employs a convolutional neural network to process time signals, the latter combines the Fast Fourier Transform operator and a convolutional neural network to analyze the VM behaviour in the frequency domain. 
In the following, we will first outline the basic elements of a deep learning approach to classification. Next, we will describe the DeepConv network and subsequently we will outline the DeepFFT model by highlighting the differences with the previous approach.

\subsection{Model overview}
We start our analysis with a discussion of the Deep learning approach to classification that is the core of the paper. 
To provide a consistent description, we will refer to the symbols and the nomenclature shown in Table \ref{table:tab1}. As further information, we provide the dimensionality/value ranges for the main elements of the model. We do not provide this information for the main layers and components of the deep-learning model (bottom part of the table) because they are described in full detail in the following of the paper.

\begin{table}[ht]
\resizebox{\columnwidth}{!}{
\begin{tabular}{|c|c|c|}
\toprule[1pt]
\textbf{Symbol} & \textbf{Meaning} & \textbf{Dimensionality}  \\
\midrule[1pt]
$M$ & Number of dataset metrics& 16\\
$W$  & Sequence length considered & 4 \dots 256 \\
$X$  & Input data               & $M \times W$ \\
$C$  & Number of VM classes     & 2 \\
$N$  & batch size               & 64 \\
$N_b$ & Number of model's blocks & $2 \dots N$ \\
$\mathcal{B}_n$ & n-th Block                 & $2 \dots N$ \\
$K_s$ & Kernel size in convolutions   & 3\\
$s$ & stride (step) in convolutions   & 2\\
$out$ & Output of the model & $C$ \\
\midrule
$\mathcal{FC}$ & Fully Connected layer    & \\ 
$\mathcal{BN}$ & Batch-Norm layer \cite{DBLP:journals/corr/IoffeS15}    & \\ 
$\mathcal{C}_{1D}$ & Convolution layer    & \\ 
$\mathcal{A}_{ReLU}$ & ReLU Activation function & \\ 
$\mathcal{B}_{FFT}$ & Fast Fourier Transform layer & \\ 
\bottomrule[1pt]
\end{tabular}
}
\caption{Summary of symbols used in the model} \label{table:tab1}
\end{table}

Convolutional Neural Networks (CNNs) are a class of Artificial Neural Networks that have been proven very effective in the last years in solving complex tasks involving multimedia data, such as images and video. 
They work well for identifying simple patterns within input data which can then be used to form more complex patterns in subsequent operations and finally be sufficiently informative to be used to perform the specific task at hand (\ie classification, regression, etc).

Inspired by deep convolutional neural networks, which are composed of several Convolutional layers followed by a final Fully Connected layer, our models are composed by a variable sequence of blocks and a final Fully Connected layer for the output classification.

VMs behavior is described as a set of time series each describing a specific metric (\eg Memory utilization, CPU utilization, Network traffic, etc\dots). For a complete list of the metric used in our experiments, the reader can refer to Table~\ref{tab:metrics} in the next section.
Since each metric is a series of samples taken across time, that are by nature one dimensional signals, we make use of a specific kind of Convolution, \ie 1-dimensional Convolutions, as elementary block of our network, and we consider each metric as an input channel on which we calculate the convolution operation independently.

A Convolution 1D is a deep learning linear operation used to extract features from one dimensional data like signals, with the aim to identify local patterns within a certain window, which is called kernel size ($K_s$). The kernel contains the learnable parameters that are used to carry out the operation. Because this kernel is being shifted along the time dimension with a certain step, called stride ($s$), the same calculation is executed on every patch of the data attended by the kernel, so that a pattern learned at one position can also be recognized at a different position, making 1D Convolution translation invariant.

In the simplest case, let's assume a Convolution 1D layer with input size $(N, C_{in}, L_{in})$ and output $(N, C_{out}, L_{out})$, then the values of the output tensor can be computed as follow:
\begin{eqnarray}
  out\left[i,j\right]=b\left[j\right] + \sum\limits_{k=0}^{C_{in} - 1} w\left[j, k\right] \star input\left[i,k\right]
\end{eqnarray}

where $\star$ is the convolution operator, $N$ is the batch size, $C$ denotes the number of channels (for the first layer, the channels correspond to the VMs metrics), $L$ is the length of signal sequence (referred to as $input$). Furthermore,  $w$ and $b$ are the learnable  parameters of the layer, with shape $(C_{out}, C_{in}, K_s)$ and $(C_{out})$, respectively.

The stride ($s$) is an hyper-parameter of the 1-dimensional convolution that controls the step of the kernel: if greater than one, data is scanned with greater steps, hence less times, causing a decrease of the initial dimension length, an effect that can be seen as pooling, a well-known deep learning strategy to reduce data dimensions, useful especially when they are high.

In our models we also make use of Batch Normalization~\cite{DBLP:journals/corr/IoffeS15}, which is a popular operation that normalizes data across the batch dimension considered at a time, where the batch is a subsample of the dataset use in training phase to speed up gradient based optimization. Applying Batch Normalization after each convolutional layer helps deep networks to converge faster; and lastly, we use the non-linear activation function ReLU (Rectified Linear Unit) that is used to stabilize the gradient during training.
\subsection{DeepConv Model}

Each block of our DeepConv network is hence comprised of a Convolution1D layer with kernel size of 3 and stride of 2, followed by a Batch-Normalization layer and a ReLU activation function; this block is repeated a variable number of times depending on the input sequence length ($W$), with a minimum of two times. Following the last block, data is then flattened and applied to a Fully Connected layer which will output the class results.

The number of blocks of the model is given by the following:
\begin{equation} 
\label{eq:N_b}
N_b = \max(\log_2(W) - 1 , 2)
\end{equation}
Where $W$ is the input sequence length of the data considered.

Varying the number of blocks is a consequence of a simple consideration: the model needs to have a final layer with neuron’s Receptive Field that can observe, and thus leverage information, over the entire input sequence; hence, given that we do experiments with different input sequence length, we need a flexible model that can adapt its architecture’s depth based on the input at hand.
The Receptive Field of a general neuron is nothing but the portion of input data that the neuron has access to and can influence its activation.

\begin{figure}
  \includegraphics[width=\linewidth]{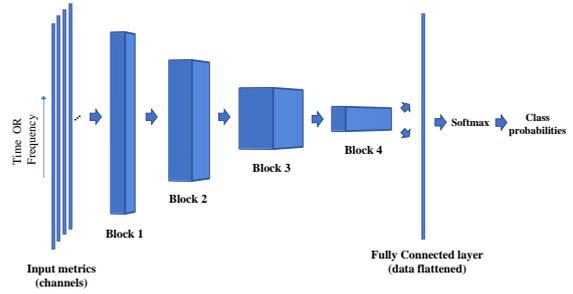}
  \caption{DeepConv model architecture.}
  \label{fig:model}
\end{figure}

Our generic DeepConv model for an input sequence length of 32 timesteps is shown in Figure~\ref{fig:model}, where we outline how the data shape changes passing through each block of the network until the final Fully Connected layer; each metric is exhibited as a column of values, so that putting together $M$ metrics (16 in our case), we obtain an input shape of ($W, M$).
Adding also the Batch-Size dimension $N$, which considers Batch-Size input samples at a time to optimize the network, we obtain the final input shape for training of ($N, W, M$).

The final $out$ class probabilities given by the model is calculated as follows:
\begin{equation} 
P(out)=\text{softmax}(out) = \frac{e^{out}}{\sum e^{out}}
\end{equation}
Where $out$ is the output of the model, which is computed as follows:
\begin{equation} 
out = (\mathcal{FC} \circ \mathcal{B}_1 \circ \mathcal{B}_2 \circ \dots \circ \mathcal{B}_{N_b})(X)
\label{eq:out}
\end{equation}
with $\circ$ being the concatenation operator for Neural Network blocks, $N_b$ the number of blocks as defined in Eq.~\ref{eq:N_b}, $\mathcal{FC}$ the final Fully Connected layer, and each block defined as:
\begin{equation} 
\mathcal{B}_n = (\mathcal{A}_{ReLU} \circ \mathcal{BN} \circ \mathcal{C}_{1D}) (X)
\label{eq:block}
\end{equation}
where $\mathcal{A}_{ReLU}$ indicates the activation function, $\mathcal{BN}$ a Batch Normalization layer, $\mathcal{C}_{1D}$ a Convolution1D layer and $X$ the input tensor.

\subsection{DeepFFT Model}

Given the periodicity of some of the signals involved, we derived a second model, namely DeepFFT, with the same architecture described so far used with time series data, but applied to data frequencies, hence transforming the data to the Fourier Domain before feeding the model. 

If we compare DeepFFT with the previously described DeepConv model, the only difference lies in the presence of an additional initial layer which computes the Fast Fourier Transform (FFT) of each metric sequence and returns the magnitude of each frequency. We empirically found that using the magnitude leads to better results than other alternatives, such as using the phase or the raw real and imaginary parts.

Therefore the substantial difference between the two models is that DeepConv works in the time domain whilst DeepFFT works in the Fourier Domain, and we can simply derive its general formulation from Eq.~\ref{eq:out} by adding an initial FFT computation block over the input (\ie $\mathcal{B}_{FFT}$). The new layer is placed before every other block, as follows:

\begin{equation} 
out = (\mathcal{FC} \circ \mathcal{B}_1 \circ \mathcal{B}_2 \circ \dots \circ \mathcal{B}_{N_b}) (\mathcal{B}_{FFT}(X))
\label{eq:outFFT}
\end{equation}

\subsection{Implementation Details}
We now discuss some details on our implementation of the DeepConv and DeepFFT models.

We implemented the models using the PyTorch framework (code is available at the following URL: \url{https://github.com/MatteoStefanini/DeepVM}).

Our implementation includes a pre-processing of the input data. Specifically, we normalize the data to have zero mean and unit variance in each channel. As the initial stream of data is partitioned into several input sequences with a length defined as the window $W$, we also leverage data augmentation techniques for sequences longer than 64 timesteps; specifically, we apply 75\% overlay between sequences, so that we can obtain more sequences for train and evaluation purpose. 

We also balance the data to have the same number of samples within each class and we split the dataset in three parts, \emph{train}, \emph{validation} and \emph{test} sets, with the chosen fractions of 0.7, 0.2 and 0.1, respectively, and then used separately in the training, validation and test phases of the models.

For the training phase we use the Cross Entropy as loss function to evaluate the predictions of the models and back-propagate the error in training phase. In all our experiments we use the Adam optimizer \cite{DBLP:journals/corr/KingmaB14}, with default values, and, after a grid search on learning rate and weight decay hyperparameters, we found that a learning rate of 0.0003 and weight decay of 0.0012 work well in most scenarios. A more detailed sensitivity analysis with multiple scenarios, however, is left as as an open issue to address as a future work. We also reduce the learning rate by a factor of 0.6 when we observe that the validation loss does not decrease for 10 consecutive epochs.

After each training epoch, which is a pass over the entire training set, we evaluate the model performance in a validation phase (using the validation fraction of the dataset). In our experiments, we train each model for 110 epochs, observing that all models converge within this range. The validation phase identifies the best performing model that is used in the final performance evaluation with the test data never used before.

%% file: 3-results.tex
\section{Experimental results}\label{sec:results}
Experiments were carried out using a dataset owned by the University of Modena and Reggio Emilia, consisting of eight real-world cloud virtual machines, monitored for a few years and divided in two classes: Web-server and SQL-server. The list of metric fed into the classifier is provided in Table~\ref{tab:metrics}. The experimental setup used is consistent with the scenario described in~\cite{Canali:TCC18}: in particular, the classes of VMs and the metrics considered are the same.

\begin{table}[ht]
\resizebox{\columnwidth}{!}{
\begin{tabular}{|c|c|}
\toprule[1pt]
\textbf{Metric} & \textbf{Description} \\
\midrule
SysCallRate & Rate of system calls [req/sec] \\
CPU & CPU utilization [\%]\\
IdleCPU & Idle CPU fraction [\%]\\
I/O buffer & Utilization of I/O buffer [\%] \\
DiskAvl & Available disk space [\%]\\
CacheMiss  & Cache miss [\%]\\      
Memory & Physical memory utilization [\%]\\
UserMem & User-space memory utilization [\%]\\
PgOutRate & Rate of memory pages swap-out [pages/sec]\\
InPktRate & Rate of network incoming packets [pkts/sec]\\
OutPktRate & Rate of network outgoing packets [pkts/sec]\\
InByteRate & Rate of network incoming traffic [KB/sec]\\
OutByteRate & Rate of network outgoing traffic [KB/sec]\\
AliveProc & Number of processes in system\\
ActiveProc & Number of active processes in run queue\\
RunTime & Execution time \\
\bottomrule[1pt]
\end{tabular}
}
\caption{Metrics used for VM classification.} \label{tab:metrics}
\end{table}

In our experiments we aim to validate the ability of the proposed model to provide an accurate identification of VMs based on their behavior. In particular, we consider our proposed models DeepConv and DeepFFT compared with other state-of-the-art solutions such as the AGATE technique~\cite{Canali:TCC18} and a PCA-based clustering solution that exploits the correlation between the time series of the VMs metrics to characterize the VMs behavior. The PCA-based clustering has been used as the best representative of traditional clustering technique in~\cite{Canali:TCC18}.
The main metric in our analyses is the \emph{accuracy}, that is the percentage of samples identified correctly by our classifier. This metric has been used consistently in previous papers on VMs identification based on clustering, such as~\cite{Canali:TCC18}. However, in some cases we refer to the dual metric, that is the classification error percentage.
As our goal is to provide a fast and accurate identification of VMs, we consider important to evaluate how the identification accuracy changes as a function of the window length $W$. Specifically, in our experiments $W$ ranges between 4 and 256 timesteps, that in minutes correspond respectively to 20 and 1280 (slightly more than 21 hours).

\begin{figure}
  \includegraphics[width=\linewidth]{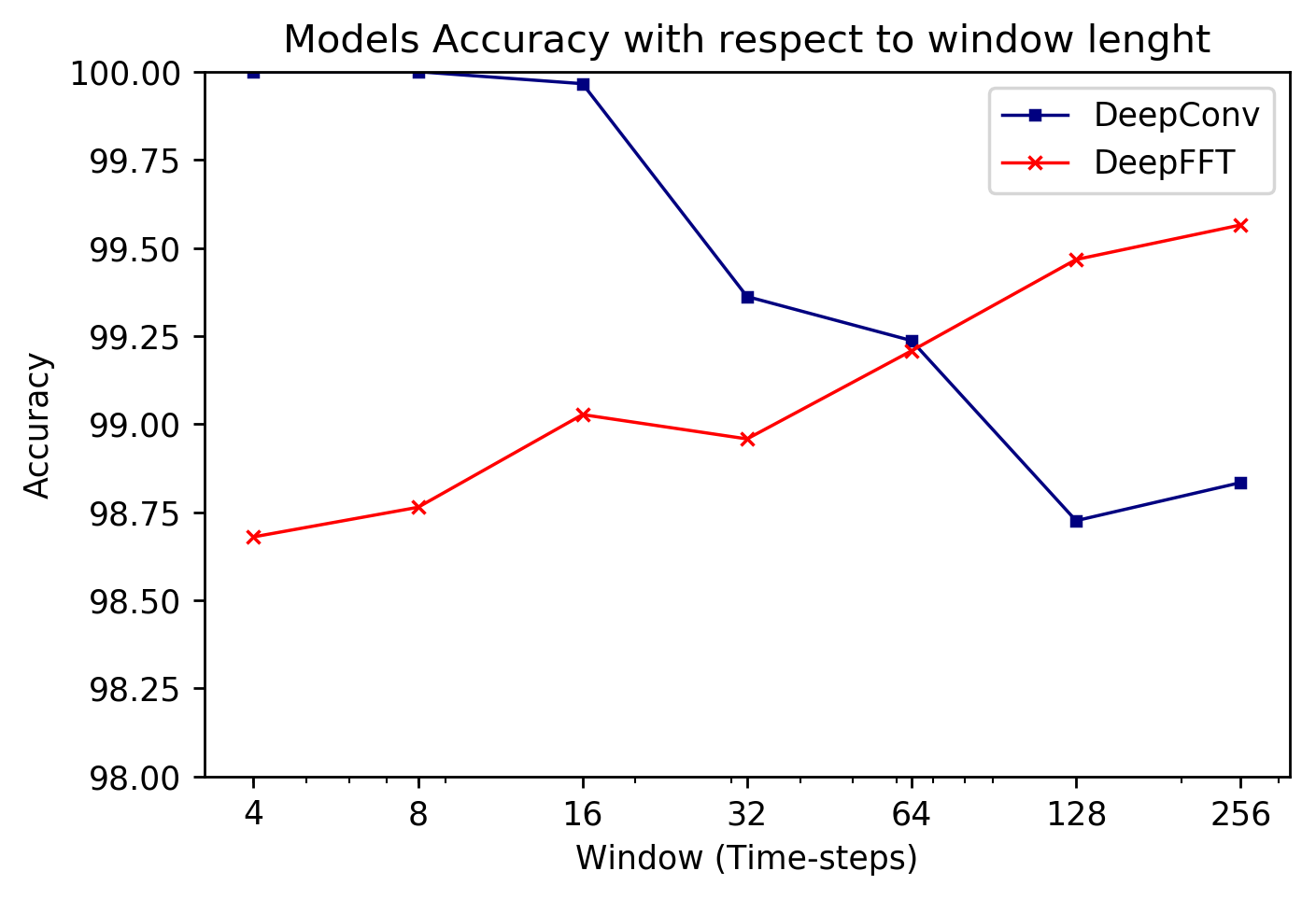}
  \caption{Models Accuracy.}
  \label{fig:accuracy}
\end{figure}

As a first result we provide an evaluation of the two proposed models, shown in Figure~\ref{fig:accuracy}. In particular, we show the accuracy achieved by the DeepConv and DeepFFT models as a function of the window length $W$ (we recall that $W$ is measured in 5 minutes time steps). A first significant result of this evaluation is the overall performance of the considered model. Looking at the data, we observe that the accuracy of each model is always higher than 98.5\%, that is a fairly good performance for this type of problems (a quick comparison with the AGATE technique, shows that both proposed models outperforms consistently AGATE). Even more interesting, the DeepConv model achieves a perfect classification of VMs, especially for short time windows, that is the most interesting and challenging scenario because it enables the identification in near real-time of a VM. The DeepConv model worsen its performance as the window increases, likely due to the increased model complexity and the fixed size of convolutional kernels. The DeepFFT model presents an opposite behavior, with performance improving as the window grows. This effect can be explained by considering that a longer time window provides more information for the Fourier-transformed problem and the classifier can work with a more consistent description of the VMs behavior.

Having obtained a first assessment of the Deep learning-based models, we compare our proposal with state-of-the-art solutions for VMs classification. In particular, we refer to the AGATE~\cite{Canali:TCC18} technique and to the PCA-based clustering.
For this comparison we refer to the error percentage and, for the AGATE technique, we also show the percentage of unclassified VMs (that are the VMs left in the gray-area for additional data collection).

\begin{figure}
  \includegraphics[width=\linewidth]{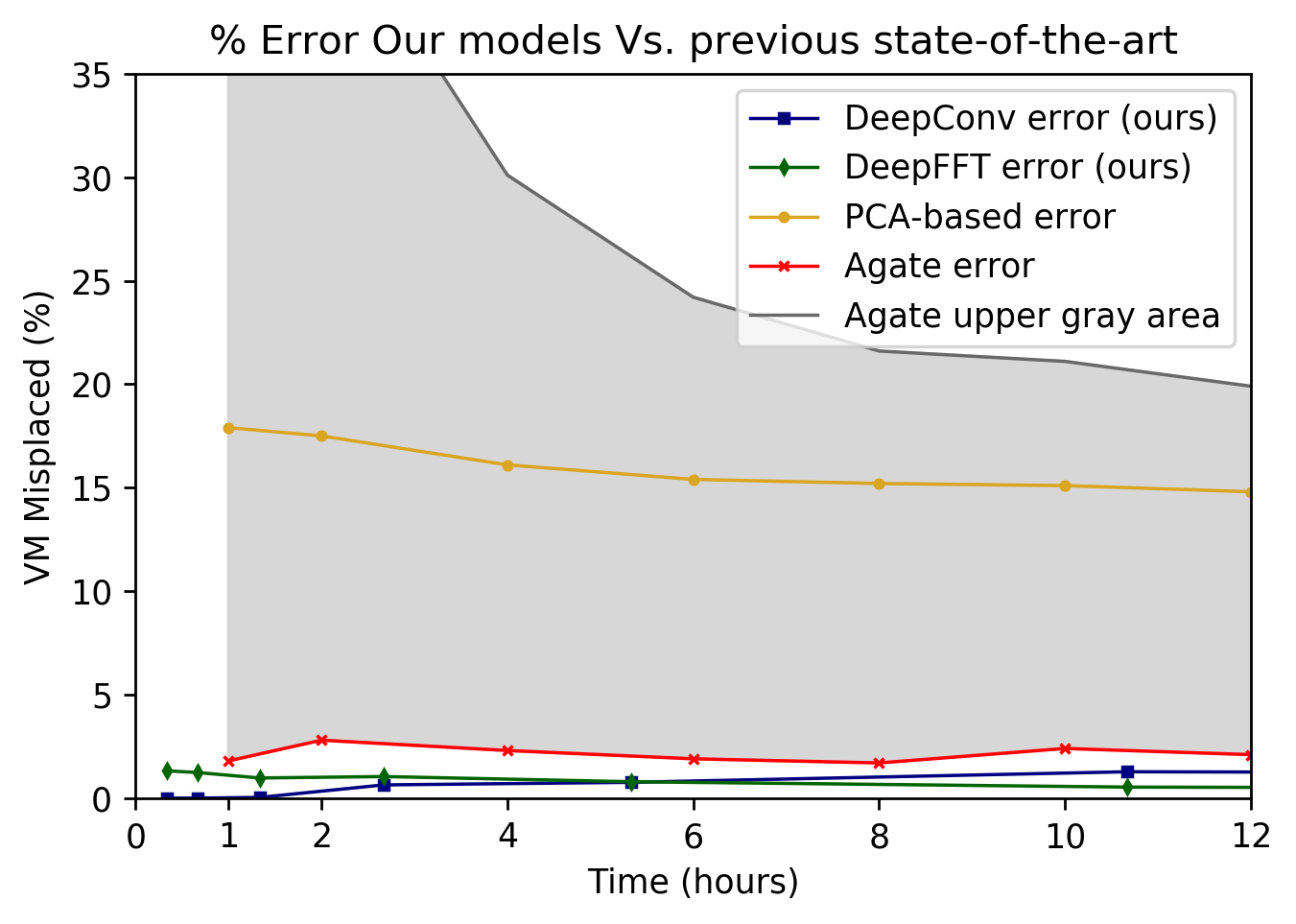}
  \caption{Comparison with state-of-the-art.}
  \label{fig:comparison}
\end{figure}

The results of the study are reported in Table~\ref{table:results} and in Figure~\ref{fig:comparison}. In particular, Table~\ref{table:results} shows the error of the considered alternatives for a window $W$ ranging from 4 to 256 time steps. As we aim to refer as closely as possible to the original results in~\cite{Canali:TCC18}, in the results marked with a star (*), we approximate the results in hours with the closest possible window size. Boldface characters are used to outline, for each value of the window $W$, the best performing solution.
We observe that, as a general rule, the longer the observation window, the better each solution is performing. For the PCA-based clustering, the percentage of errors decreases from nearly 18\% to roughly 15\%. For the AGATE solution the error percentage is not monotone but remains in the range $[1.8\%, 2.4\%]$; however we observe a clear reduction of the un-classified VMs dropping from nearly 50\% to nearly 19\% as the window grows. The AGATE solution outperforms by one order of magnitude the previous solutions. However, the Deep learning-based models are a clear step ahead compared to the AGATE technique. We observe that, for every considered window the best performance are achieved by a deep learning-based model. Furthermore, for small windows (\ie $W \le 8$ time steps), the DeepConv model achieves 0\% errors with no need to postpone any classification using the notion of a gray-area.

\begin{table}[ht]
\resizebox{\columnwidth}{!}{
\begin{tabular}{|c c|c|c|c|c|c|c|c|}
\toprule[1pt]
\multicolumn{2}{|c|}{\textbf{Method}} & \textbf{4} & \textbf{8} & \textbf{16} & \textbf{32} & \textbf{64} & \textbf{128} & \textbf{256} \\
\midrule[1pt]
\multicolumn{2}{|c|}{PCA-based} & - & 17.9* & 17.5* & 16.1* & 15.4* & 15.2* & 15.1* \\
\midrule[1pt]
\multirow{2}{*}{AGATE} & error  & - & 1.8* & 2.8* & 2.3* & 1.9* & 1.7* & 2.4* \\
 & grey-area  & - & 47.8* & 41.1* & 27.8* & 22.3* & 19.9* & 18.7* \\
\midrule[1pt]
\multicolumn{2}{|c|}{DeepFFT} & 1.32 & 1.24 & 1.45 & 1.04 & 0.79 & \textbf{0.53} & \textbf{0.43} \\
\midrule[1pt]
\multicolumn{2}{|c|}{DeepConv}& \textbf{0.00} & \textbf{0.00} & \textbf{0.03} & \textbf{0.64} & \textbf{0.76} & 1.27 & 1.17  \\
\bottomrule[1pt]
\end{tabular}
}
\caption{Error comparison with state-of-the-art.} \label{table:results}
\end{table}

The results in Table~\ref{table:results} are more clearly visible if we refer to Fig.~\ref{fig:comparison}. The reduction of the gray area (shown as a gray shadow in the figure) is quite evident and demonstrates how the AGATE technique becomes more effective with time, while the amount of errors (red line) remains in the order of few percentage points. On the other hand, the previously proposed PCA-based clustering (the yellow line) is clearly affected by an unacceptable amount of errors. However, the two deep learning-based models (blue and green lines) are a clear step ahead compared to the existing techniques as they provide even lower error rate than the AGATE alternative, without using a gray area.

%% file: 4-related.tex
\section{Related work}\label{sec:related}
The tasks of monitoring and managing Cloud data centers in a scalable way has received lots of attention over the last years~\cite{Aceto:2013,Belogazov:TSE}.

At the level of monitoring scalability, it is common to exploit aggregation and filtering techniques to reduce the amount of data before sending them to the data center controller. 

When focusing on the monitoring of a Cloud infrastructure, scalability problems are typically addressed relying on some form of dimensionality reduction (\eg filtering or aggregation) that occurs before sending data to the cloud management function. Such dimensionality reduction is performed by ad-hoc software, typically in the form of a library or implemented as an data-collecting software agent. Example of this approach are provided in~\cite{Mehrotra:2011,Shao:2011,Azmandian:2011,Kertesz:2013}.
The actual aggregation policy may vary ranging from extraction of high-level performance indicators~\cite{Shao:2011}; to obtaining parameters that aggregate metrics from different system layers (hardware, OS, application and user) using Kalman filters~\cite{Mehrotra:2011}; to a simple linear combination of OS-layer metrics~\cite{Azmandian:2011}; up to systems that extract data from both the OS and the applications~\cite{Kertesz:2013,Andreolini:CIT11}.

The problem of cloud monitoring is addressed not just by research proposals but also by full-featured frameworks both commercial or open source (just to give a few names, Amazon cloud Watch is a commercial product, while MONASCA, the OpenStack monitor is open source) 
However, the common limit of these solution is that each object taken into account in the monitoring process (either VM or physical node) is considered as independent form every other objects. In doing so, these proposals fail to take advantage form the similarities between objects exhibiting a similar behavior.

The management of Cloud systems is another critical topic, where several papers have been published, starting from the early examples based on the principles of autonomic computing applied to the the Cloud~\cite{Buyya2012}. Another interesting example of Cloud managment is represented by the Bobtail library~\cite{Bobtail}, that aims at supporting in each VM the identification of placement problems that result in high communication latency. 
All these solutions rely on the assumption that the cloud user is willing to install a specific software layer on each VM, to overcome some limitations of the IaaS vision of the cloud. Our focus is completely different as we place no requirement on the VMs user and we comply completely with the IaaS vision.
Other studies aiming at improving the data center scalability have been proposed, such as~\cite{Canali:TGCN,Canali:NCCA15,Meo:PCV}. Our proposal can be integrated with these solution to improve the scalability of the cloud data center management.

Identifying similarities between VMs in a Cloud infrastructure is the key problem of our research. Several relevant works are discussed in the following. The research in~\cite{Zhang:IOT} aims at identifying similar VMs, but the similarity detection is limited to storage resources and its application scope is that of storage consolidation strategies. Similarly, the study in~\cite{Jayaram:2011} investigates similarities of VMs static images used in public cloud environments to provide insights for de-duplication and image-level cache management. Our approach focuses on a wider range of applications because we do not limit our analysis to a few resources for a limited purpose, but we consider a robust and general-purpose multi-resource similarity identification mechanism. A similar focus on similarity detection in VMs characterizes~\cite{Canali:TCC18}. Such study aims to address the trade-off between a fast identification of VMs and its accuracy using an adaptive approach. Our proposal addresses the same issue relying on a deep learning approach that ensures a very fast and accurate identification of the VMs.

Techniques derived from deep-learning have been recently proposed to address problems in the field of distributed infrastructure such as Cloud data centers and Fog systems. For example, the authors of~\cite{Liu:HFC17} propose a deep reinforced learning technique for the management of VMs allocation in Cloud data center. 
Our proposal is completely orthogonal to the proposal in~\cite{Canali:NCCA15} and can be integrated with a class-based approach leveraging the VMs identification proposed in this paper. Another application of Deep-learning in distributed system is related to anomaly or attack detection. For example,~\cite{Diro:DAD18} proposes a deep-learning classifier for attack detection in a Fog system. While the basis of their deep-learning approach are similar to the ones in our proposal, to the best of our knowledge, ours is the first attempt to use deep learning to classify the behavior of VMs to support monitoring or management purposes, rather than aiming to attack or anomaly detection.


%% file: 5-conclusions.tex
\section{Conclusions and future work}\label{sec:conclusions}
In this paper we focused on the scalability problems of a Cloud infrastructure, aiming to enable the adoption of solutions that improve scalability of monitoring and management through a classification of VMs that exhibit a similar behavior.

Existing solution for VMs clustering and classification are characterized by a trade-off between accurate VMs identification and timely response. Previous proposals aiming to address this problem exploited the notion of a gray area. While this approach is viable for the identification of VMs with a long life span, it is hard to apply in cloud infrastructures with on-demand VMs that are typically created and destroyed in just a few hours.

This limitation motivates our proposal of a different approach to the problem that, instead of an unsupervised clustering, exploits classifiers based on deep learning techniques to assign a newly deployed VMs to a cluster of already-known VMs. We propose two deep learning models for the classifier, namely DeepConv and DeepFFT based on convolution neural networks and Fast Fourirer Transform.

We validate our proposal using traces from a real cloud data center and we compare our classifiers with state-of-the-art solutions such as the AGATE technique (that exploits a gray area to adapt the observation time of each VM so that uncertainly classified VMs are not immediately assigned to a group) and a PCA-based clustering solution.
The results confirm that the deep learning models consistently outperforms every other alternative without the need to introduce a gray area to delay the classification. Even more interesting, the proposed classifiers can provide a fast and accurate identification of VMs. In particular, the DeepConv model provides a perfect classification  with just a 4 samples of data (corresponding to 20 minutes of observation), making our proposal viable also to classify on-demand VMs that are characterized by a very short life span.

This paper is just a preliminary work in a new line of research that aims to apply deep learning techniques to the problems of cloud monitoring and management. Future works will focus on a more thorough evaluation of the proposed models, with additional sensititiy analyses with respect to the models parameters; on the proposal of additional classification models; and on the application of Generative Adversarial Networks to improve the quality of VMs identification in cases where the quality of data is lower than in the considered example (i.e., due to reduced number of metrics and presence of sampling errors).